\documentclass[10pt,twocolumn,letterpaper]{article}

\usepackage[pagenumbers]{cvpr} 

\usepackage{booktabs}
\usepackage{threeparttable}
\usepackage{multirow}
\usepackage{colortbl}

\definecolor{cvprblue}{rgb}{0.21,0.49,0.74}
\usepackage[pagebackref,breaklinks,colorlinks,allcolors=cvprblue]{hyperref}

\title{Revealing Key Details to See Differences: A Novel Prototypical Perspective for Skeleton-based Action Recognition}

\author{
Hongda Liu\textsuperscript{1,2}, Yunfan Liu\textsuperscript{2}$^{\ast}$, Min Ren\textsuperscript{3}, Hao Wang\textsuperscript{1,2}, Yunlong Wang\textsuperscript{1}, Zhenan Sun\textsuperscript{1}\thanks{Corresponding authors.}\\
\textsuperscript{1}NLPR, Institute of Automation, Chinese Academy of Sciences\\
\textsuperscript{2}University of Chinese Academy of Sciences\quad
\textsuperscript{3}Beijing Normal University\\
{\tt\small \{hongda.liu,yunfan.liu,hao.wang,yunlong.wang\}@cripac.ia.ac.cn}\\
{\tt\small renmin@bnu.edu.cn, znsun@nlpr.ia.ac.cn}
}

\begin{document}
\maketitle

\begin{abstract}
In skeleton-based action recognition, a key challenge is distinguishing between actions with similar trajectories of joints due to the lack of image-level details in skeletal representations.
Recognizing that the differentiation of similar actions relies on subtle motion details in specific body parts, we direct our approach to focus on the fine-grained motion of local skeleton components.
To this end, we introduce ProtoGCN, a Graph Convolutional Network (GCN)-based model that breaks down the dynamics of entire skeleton sequences into a combination of learnable prototypes representing core motion patterns of action units.
By contrasting the reconstruction of prototypes, ProtoGCN can effectively identify and enhance the discriminative representation of similar actions.
Without bells and whistles, ProtoGCN achieves state-of-the-art performance on multiple benchmark datasets, including NTU RGB+D, NTU RGB+D 120, Kinetics-Skeleton, and FineGYM, which demonstrates the effectiveness of the proposed method.
The code is available at \url{https://github.com/firework8/ProtoGCN}.
\end{abstract}

\section{Introduction}

Human action recognition, a pivotal task with various real-world applications, has recently seen growing interest in skeleton-based methods due to their efficiency and robustness.
Given the intrinsic graph structure of the human skeleton, Graph Convolutional Networks (GCNs)~\cite{yan2018spatial,shi2019two,chen2021channel,chi2022infogcn,lee2023hierarchically} are widely used to model the relationships between joints.
However, despite their effectiveness in action recognition, these methods struggle to distinguish actions with similar motion patterns.

\begin{figure}[t]
\centering
\includegraphics[width=\linewidth]{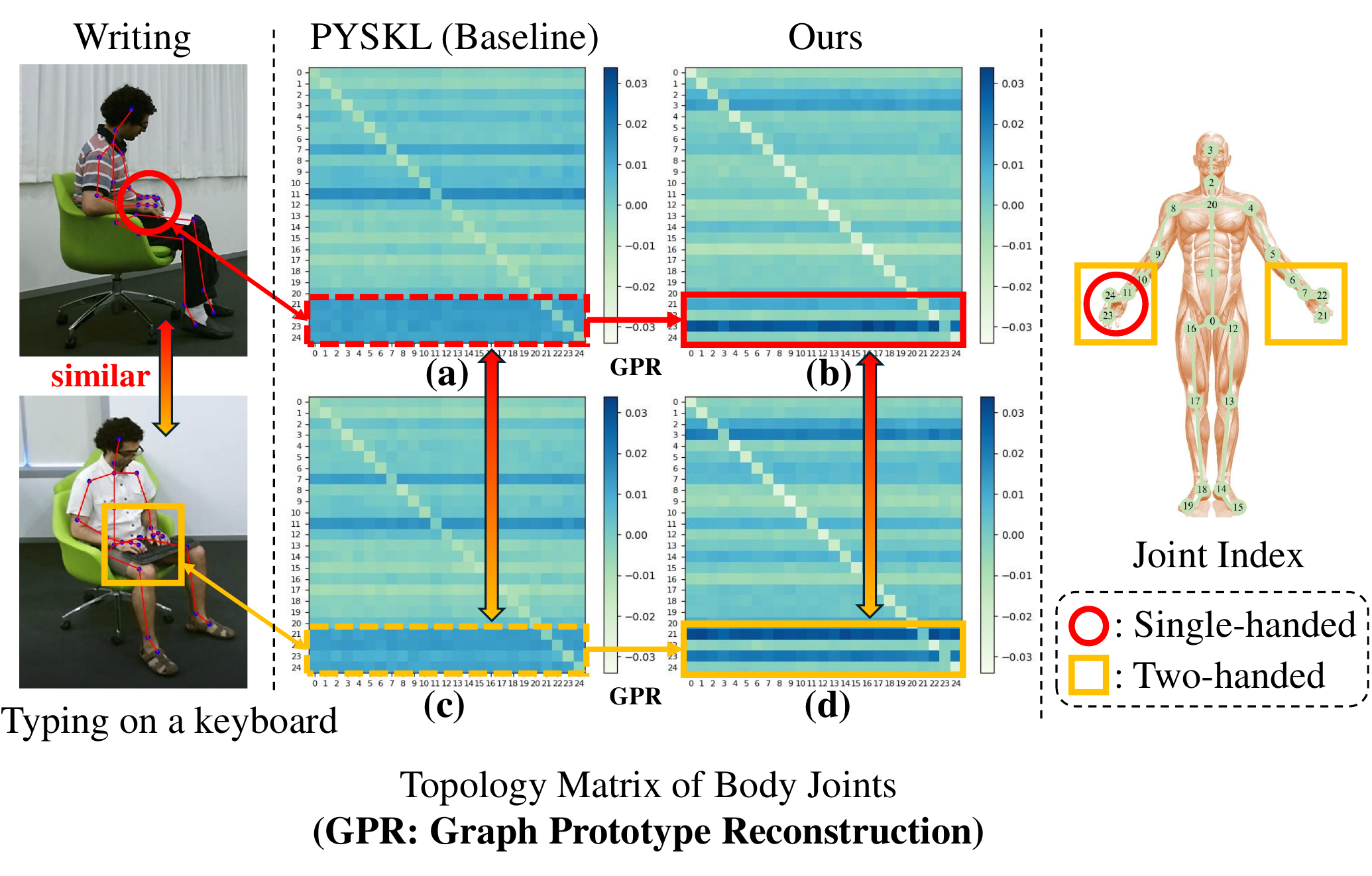}
\setlength{\abovecaptionskip}{-8pt}
\setlength{\belowcaptionskip}{0pt}
\caption{ 
Illustration of the skeletons and learned topologies for similar actions \textit{Writing} and \textit{Typing on a Keyboard} (deeper color indicates stronger relationships between corresponding joints).
As shown in (a) and (c), the baseline PYSKL~\cite{duan2022pyskl} demonstrates its ability to focus on joints associated with hands, but falls short in revealing their distinctive motion characteristics.
In contrast, the integration of the Graph Prototype Reconstruction mechanism facilitates a clearer differentiation between the two actions, as evidenced by the notably distinct motion patterns observed between (b) and (d).
Please zoom in for a better view.
}
\label{fig:figure1}
\vspace{2pt}
\end{figure}

Specifically, existing methods face a key challenge in capturing the fine-grained details of crucial body parts, which are essential for distinguishing similar actions.
Skeleton-based action recognition is primarily concerned with two tasks: identifying the key body parts involved and understanding their motion patterns.
However, due to the lack of image-level information, current methods tend to capture coarse-grained motion patterns, effectively identifying the key parts but struggling to capture the subtle, discriminative details that are crucial for distinguishing similar actions.
This limitation arises because the patterns of similar actions often overlap significantly, with distinctive details representing only a small portion of the overall information, making it difficult for the classifier to differentiate between them.

As illustrated by the graph topologies in \cref{fig:figure1}, PYSKL~\cite{duan2022pyskl} successfully focuses on the hand-related joints when distinguishing between \textit{writing} and \textit{typing on a keyboard}.
However, the high similarity between the graph topologies reveals its challenge in capturing the subtle distinctions in the motion characteristics of the attended joints.
Specifically, in \textit{typing on a keyboard}, both hands exhibit more balanced movement, whereas in \textit{writing}, one hand is more actively engaged.
Unfortunately, this difference is not well captured in the learned topologies shown in \cref{fig:figure1}.
Furthermore, intra-class variations, such as changes in viewpoint or motion amplitude, inevitably introduce noise into the acquired motion patterns. 
As a result, the task of capturing representative and discriminative motion features becomes more challenging, hindering the accurate classification of similar actions.

The challenge outlined above has driven us to seek a representation that more effectively captures the intrinsic characteristics of actions with high discriminative power. 
To address this, we propose ProtoGCN, a novel graph prototype learning approach for skeleton-based action recognition.
At the heart of ProtoGCN lies the Prototype Reconstruction Network, which explicitly enforces the representation to be formed as an associative combination of action prototypes stored in a memory module.

In specific, these prototypes represent diverse relationship patterns between all the human joints.
The network samples and assembles the relevant prototypes via the associative response of input action features, producing the uniquely tailored representations.
In addition, a motion topology enhancement module is incorporated to provide richer and more expressive features, facilitating prototype retrieval.
Furthermore, we propose a class-specific contrastive learning strategy to accentuate the inter-class distinction, which compels the model to uncover distinctive prototypes to associate all action classes in the dataset.
Such a prototype-based associative reconstruction process helps to enhance the discriminative capability of the resultant motion features and eliminate irrelevant distracting information.
Both qualitative and quantitative analyses showcase the superior efficacy of ProtoGCN in acquiring discriminative representations and discerning similar actions.

The main contributions are summarized as follows:
\begin{itemize}
  \item We introduce ProtoGCN, a novel graph prototype learning method that amplifies subtle distinctions for accurate differentiation of similar actions.
  
  \item A Prototype Reconstruction Network is proposed to adaptively assemble prototypes, yielding distinctive representations under a contrastive objective.
  
  \item Extensive experiments on four large-scale benchmarks (NTU RGB+D, NTU RGB+D 120, Kinetics-Skeleton, and FineGYM) demonstrate that ProtoGCN consistently achieves state-of-the-art performance.
\end{itemize}

\section{Related Work}

\subsection{Skeleton-based Action Recognition with GCNs}

Considering the intrinsic graph-like structure of human skeletons, the majority of GCN-based approaches~\cite{yan2018spatial,shi2019two,liu2020disentangling,chen2021multi,chen2021channel,duan2022dg,zhou2024blockgcn} explicitly model the relationships using spatial-temporal graphs.
ST-GCN~\cite{yan2018spatial} is the pioneering work integrating graph convolutions for comprehensive spatial-temporal modeling.
Most subsequent studies adopt the main framework of ST-GCN and place greater emphasis on enhancing the capacity of GCNs. 
2s-AGCN~\cite{shi2019two} proposes an adaptive GCN model to learn topological correlations.
Following that, CTR-GCN~\cite{chen2021channel} introduces the channel-wise graph convolution to explore topology-non-shared graphs.
Similarly, InfoGCN~\cite{chi2022infogcn} models context-dependent intrinsic topology, simultaneously leveraging an information-theoretic objective to represent latent information.
HD-GCN~\cite{lee2023hierarchically} proposes the hierarchically decomposed graph to model distant spatial relationships.

Tracing the evolution of GCN-based approaches, the learning of graph topologies in GCNs plays a crucial role.
Nevertheless, existing methods predominantly focus on general structural discrepancies in motion patterns, while frequently neglecting the nuanced details inherent in the motion features of specific key body parts.
In contrast, we propose a novel graph prototype learning approach aimed at amplifying critical detail information and capturing nuanced motion patterns.

\subsection{Prototype Learning}

Traditionally, prototype learning is a classical method that establishes a metric space and optimizes the most representative anchors from training data~\cite{lee2023unsupervised}.
Due to the exemplar-driven nature and simpler inductive bias, prototype learning has recently shown significant potential in a variety of tasks~\cite{snell2017prototypical,yang2018robust,wang2019panet,ramsauer2020hopfield,yang2020prototype}.
Interestingly, human vision itself deploys a remarkable prototyping ability, skillfully focusing on principal body joints while filtering out extraneous elements~\cite{simon1971human,giese2003neural}.
In this context, we argue that movements of human body joints likewise exhibit noteworthy similarities, resulting in a collective of prototypes.
Integrating prototype learning into the network facilitates the natural encapsulation of diverse dynamic characteristics, enhancing the capability to discern similar actions.

Therefore, in this study, we employ prototype learning for diverse motion patterns through the utilization of the Prototype Reconstruction Network.
Due to the considerable difficulty in directly learning highly detailed features for various actions, we represent them as combinations of learnable prototypes to highlight key information.
The proposed method discovers and assembles the most representative prototypes, which can subsequently be employed to discern more fine-grained and distinct semantics.

\subsection{Contrastive Learning}

Recently, contrastive learning has been widely applied in self-supervised learning studies~\cite{bachman2019learning,tian2020contrastive,chen2020improved}.
Representative methods such as MoCo~\cite{he2020momentum} and SimCLR~\cite{chen2020simple}, use the InfoNCE loss to pull the positive samples close while pushing them away from the negative samples in the embedding space. 
In skeleton-based action recognition, previous contrastive learning-based approaches~\cite{li20213d,guo2022contrastive,shah2023halp,zhang2023prompted} concentrate on directly obtaining accurate skeleton representations.
Under the supervised setting, these approaches~\cite{huang2023graph,zhou2023learning,chang2024wavelet} mainly focus on comparing the learned features to further optimize the model. 
SkeletonGCL~\cite{huang2023graph} explicitly explores the global context across all sequences through contrastive learning.
FR-Head~\cite{zhou2023learning} proposes contrastive feature refinement to obtain discriminative representations of skeletons.
In contrast, we adopt the contrastive training objective as a form of regularization, urging the reconstruction mechanism to identify and distinguish the distinctive prototypes.

\section{Method}

\begin{figure*}[t]
\centering
\includegraphics[width=\linewidth]{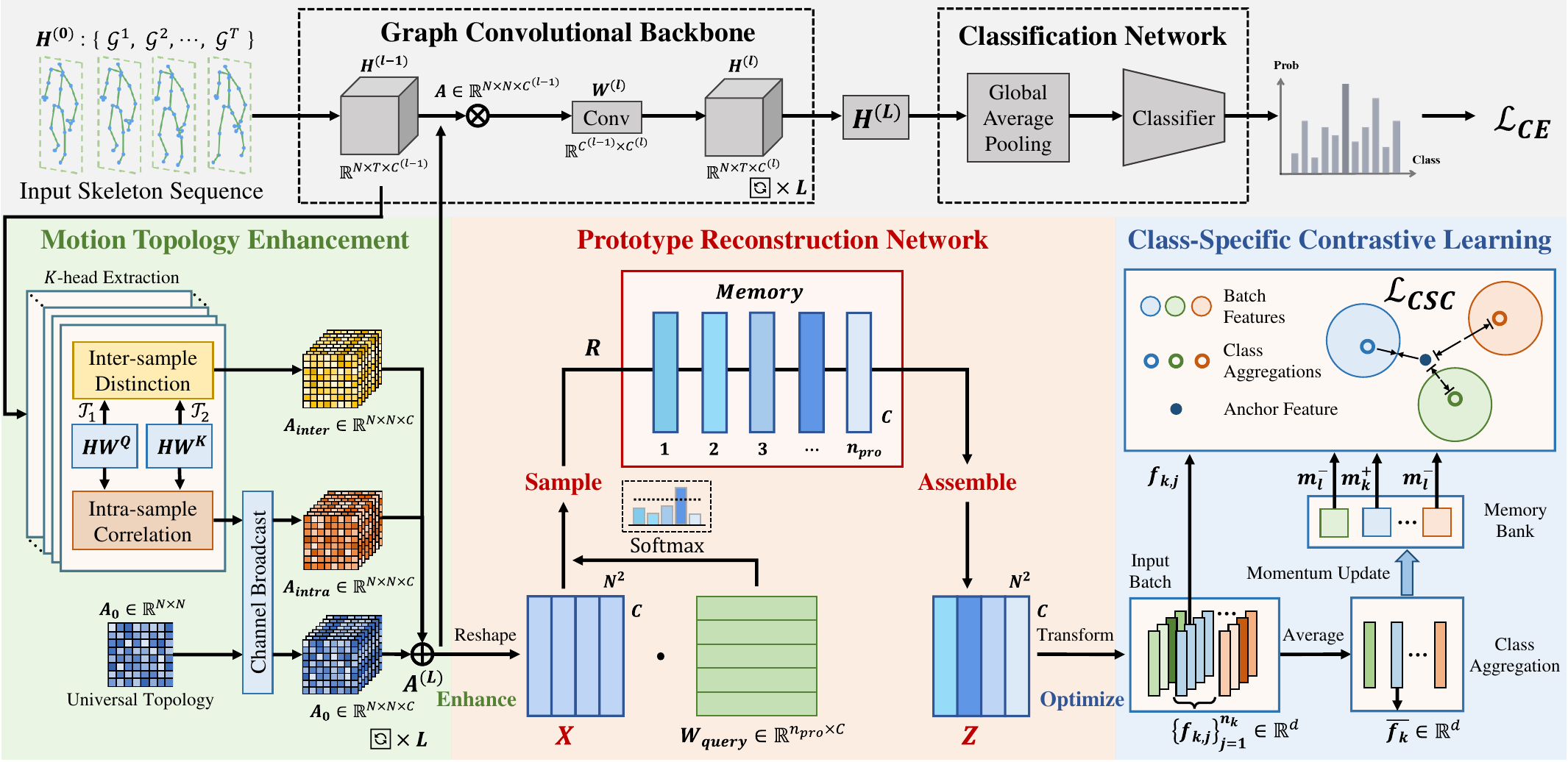}
\setlength{\abovecaptionskip}{0pt}
\setlength{\belowcaptionskip}{0pt}
\caption{ 
The overall architecture of ProtoGCN. 
A Prototype Reconstruction Network is proposed to transform the representation of graph topology $\mathbf{X}$ into a combination $\mathbf{Z}$ of learnable prototypes at the fine-grained joint level, thereby enhancing the distinctiveness of features. 
In specific, the prototypes represent diverse relationship patterns between all the human joints. 
Additionally, at each layer of the network, the Motion Topology Enhancement module is integrated to capture rich and expressive motion representations, establishing the foundation for prototype learning. 
Last, the outputs of the model are supervised by the classification loss and class-specific contrastive loss, respectively. 
}
\label{fig:figure2}
\vspace{-1mm}
\end{figure*}

In this section, we first briefly introduce the preliminary concepts related to skeleton-based action recognition with GCNs, then provide a detailed description of the main components of ProtoGCN.
An overview of the framework of ProtoGCN is presented in \cref{fig:figure2}.

\subsection{Preliminaries}
\label{sec:3_1}

Due to the intrinsic structural nature, the human skeleton is commonly represented as a graph $\mathcal{G}=(\mathcal{V},\mathcal{E})$, where $\mathcal{V}$ and $\mathcal{E}$ denote the joint and edge sets.
The input skeleton sequence is represented as a feature tensor $\mathbf{H}\in\mathbb{R}^{N\times T\times C}$, where $N$, $T$, and $C$ are the number of body joints, frames, and feature dimension, respectively.
Generally, GCN models stacked graph convolutional layers ($L$ layers in total) to extract the semantic feature of skeletons.

For current adaptive methods, the translation function between the hidden variables in the $(l-1)$\textsuperscript{th} and $(l)$\textsuperscript{th} layer, \ie, $\mathbf{H}^{(l-1)}$ and $\mathbf{H}^{(l)}$, can be formulated as 
\begin{equation}
  \mathbf{H}^{(l)} = \sigma (\mathbf{A}^{(l)}\mathbf{H}^{(l-1)}\mathbf{W}^{(l)})
  \label{eq_1}
\end{equation}
where $\mathbf{A}^{(l)}\in\mathbb{R}^{N\times N\times C}$ denotes the adaptive graph topology matrix, $\mathbf{W}^{(l)} \in \mathbb{R}^{C^{(l-1)}\times C^{(l)}}$ represents the learnable weight matrix utilized for feature projection, and $\sigma$ is the ReLU activation function~\footnote[1]{The layer index $l$ will be omitted in the remainder of this paper for conciseness unless its specification is necessary.}.
After obtaining the final hidden representation $\mathbf{H}^{(L)}$, a classification network is employed to determine the prediction label $\hat{\mathbf{y}}\in \mathbb{R}^c$ ($c$ denotes the total number of action classes).
Finally, a cross-entropy loss $\mathcal{L}_{CE}$ is used to supervise the predicted action class
\begin{equation}
  \mathcal{L}_{CE} = -\sum_{i}^{c} \mathbf{y_{i}} \log \hat{\mathbf{y_{i}}} 
  \label{eq_2}
\end{equation}
where $\mathbf{y_{i}}$ is the one-hot ground-truth action label.
Following this routine, the approach adopted by ProtoGCN will be elaborated in the subsequent subsections.

\subsection{Prototype Reconstruction Network}\label{sec:3_2}

As previously discussed, the essence of distinguishing similar actions lies in recognizing the variations in the motion patterns of involved body joints. 
However, effectively capturing these differences to create a discriminative representation poses a significant challenge. 
The nuanced nature of the corresponding joint movements often leads to these distinctions becoming obscured within the total body motion.

To tackle this issue, the model's attention needs to be directed towards identifying the detailed joint-level motion characteristics.
Therefore, we propose the Prototype Reconstruction Network (PRN) to obtain the fine-grained enhanced representations.
This is achieved by decomposing the representations into a combination of the learned prototype vectors to highlight key details.

Specifically, the network consists of two parts: a memory module and a reconstruction module.
The memory module stores the learned action prototypes that represent diverse relationship patterns between human body joints.
Accordingly, the reconstruction module computes the response weights of input representations through an addressing mechanism.
Last, the network samples and assembles the relevant prototypes via the response, producing uniquely tailored feature representations.

\noindent \textbf{Memory Module.}
Given the input skeleton sequences, the learned graph representation, \ie, $\mathbf{A}\in\mathbb{R}^{N\times N\times C}$, is first reshaped to $\mathbf{X}\in\mathbb{R}^{N^{2}\times C}$.
The representation $\mathbf{X}$ can be intuitively interpreted as a feature aggregation that represents the correlations between joints.
Then we implement memory as a learnable matrix $\mathbf{W_{memory}} \in\mathbb{R}^{n_{pro}\times C}$.
The prototype vector $m_{i}$ (for $i \in \{ 1, \ldots, n_{pro}\}$) in each row of the matrix represents the learned relationship patterns between joints, where $n_{pro}$ denotes the number of prototypes.

In practice, $\mathbf{W_{memory}}$ is randomly initialized and the stored patterns are adaptively updated throughout training.
Instead of fixed iterative updates, we make the network automatically exploit the clusters of these relationships through backward gradient, and learn the representative prototypes from the input graph embeddings.
These learned prototypes can then be composed into unique enhanced representations under the response signals to be introduced below.

\noindent \textbf{Reconstruction Module.}
Our goal is to leverage the relevant prototype patterns to enhance the input feature, yielding more significant and discriminative representations.
We use attention as the addressing mechanism.
For the input graph representation, the module produces a response signal that represents a uniquely tailored linear combination of prototypes in the memory module.

Concretely, a learnable matrix $\mathbf{W_{query}} \in \mathbb{R}^{n_{pro}\times C}$ serves as the key vectors projected from the input features, and then a softmax activation function is employed to compute the response signals.
The addressing process could be further expressed as
\begin{equation}
  \mathbf{R} = {\rm softmax} (\mathbf{X}{(\mathbf{W_{query}})}^\top)
  \label{eq_3}
\end{equation}
where the signals $\mathbf{R}$ are used as weights to combine the prototypes.
Here the softmax provides a weighted average of the target vectors based on the similarity between the query and the input.
Then we can accomplish memory activation and obtain an enhanced output representation: 
\begin{equation}
  \mathbf{Z} = \mathbf{R} \cdot \mathbf{W_{memory}}
  \label{eq_4}
\end{equation}

The entire network could be considered as an encoder-decoder bottleneck model, where the memory is filled with the encoder output set.
It is trained in an end-to-end supervised manner, with the memory updated while training along with the rest of the network parameters.
Instead of relying on a pre-determined biological decomposition of the human skeleton, we enable the model to adaptively learn the composition of human actions.
Given an input representation, the network generates tailored latent responses and then reconstructs the unique enhanced representation, \ie, $\mathbf{X}$ to $\mathbf{Z}$.
An illustration of this `sample-and-assemble' process is provided in \cref{fig:figure2}.

\subsection{Motion Topology Enhancement}
\label{sec:3_3}

The quality of input topology representations $\mathbf{X}$ is vital since they determine the retrieval efficiency of the learned prototypes.
Therefore, we present the Motion Topology Enhancement (MTE) module to augment the capacity to capture intricate motion details, resulting in the abundant and expressive topologies.

In practice, we start by mapping the input feature of each layer $\mathbf{H}\in\mathbb{R}^{N\times T\times C}$ to $K$ sets of queries and keys with learnable matrices $\mathbf{W^Q}, \mathbf{W^K} \in \mathbb{R}^{C\times C'}$.
The attention mechanism effectively establishes the global perceptual connections between different body joints.
Multi-head configuration and mapping functions could further facilitate the diverse feature semantic with the dimension $C'=[{C}/{K}]$.
Subsequently, we adopt average pooling to perform the aggregation of contextual temporal relations and squeeze the dimension $T$.
The projected features could be formulated as $H^Q = \mathbf{H}\mathbf{W}^{Q}\in \mathbb{R}^{N\times C'}$ and $H^K = \mathbf{H}\mathbf{W}^{K}\in \mathbb{R}^{N\times C'}$.

Then we model the nuanced relationships, which consist of two parts, the intra-sample correlations and the inter-sample distinctions.
In delineating the details, intra-sample correlations exhibit skeleton-level comprehensive connections, requiring self-attention to identify their specific patterns.
Moreover, for samples with inconsistent motion patterns, inter-sample distinctions could be captured by pairwise comparisons between joints.
The differential realization eliminates extraneous noise, encouraging models to focus on critical information.

Specifically, the interdependence of intra-sample correlations is modeled by the inner product of projected features.
For comparing distinctions, we adopt the transformation $\mathcal{T}_d(\cdot)$ to expand the tensor at its $d$-th dimension and replicate the result $N$ times along that dimension.
As a result, $\mathcal{T}_1 (H^Q)$ and $\mathcal{T}_2 (H^K)$ have the same dimensions of $\mathbb{R}^{N\times N\times C'}$. 
The enhanced topology representations could be efficiently computed by
\begin{align}
\mathbf{A_{intra}} &= {\phi}_{<intra>} (H^Q(H^K)^\top) \label{eq_5} \\
\mathbf{A_{inter}} &= {\phi}_{<inter>} (\mathcal{T}_1 (H^Q)-\mathcal{T}_2 (H^K)) \label{eq_6}
\end{align}
where $\phi$ represents the activation function.
Then these representations are integrated into each GCN layer of ProtoGCN.
They rely on specific input skeletons and delineate nuanced details of joints relevant to the corresponding action, thereby serving as a crucial component in distinguishing similar actions.
In contrast, the original dynamic topology $\mathbf{A}$ is represented as $\mathbf{A_0}$, which is utilized to represent the universal interconnections among human body joints and shared across all action classes.

Substituting the above decomposition of the graph topology into \cref{eq_1} gives the following translation function.
\begin{equation}
  \mathbf{H}^{(l)} = \sigma ((\mathbf{A_0}+\mathbf{A_{intra}}+\mathbf{A_{inter}})\mathbf{H}^{(l-1)}\mathbf{W}^{(l)})
  \label{eq_7}
\end{equation}
In this way, the network incorporates nuanced motion intricacies of the acquired topology, establishing the foundation for prototype learning.

\subsection{Class-Specific Contrastive Learning} 
\label{sec:3_4}

Basically, in \cref{eq_4}, $\mathbf{Z}$ could be regarded as the weighted sum of prototypes retrieved from $\mathbf{W_{memory}}$, where the weights are determined by the input-dependent probe $\mathbf{R}$.
However, the formulation itself does not ensure the representativeness of learned prototypes or the discrimination of the resultant feature $\mathbf{Z}$, and thus it may not necessarily lead to improved performance in classifying similar actions.

Therefore, to further accentuate the inter-class distinction of human actions, the class-specific contrastive learning strategy is incorporated.
Specifically, the strategy introduces two components: a compactness constraint that encourages samples to be close to their class-specific aggregations, and a dispersion contrastive loss that promotes the distinctions among different class aggregations.

In practice, a projection network is first utilized to embed $\mathbf{Z}$ into vectors within the contrastive feature space.
We compress $\mathbf{Z}\in\mathbb{R}^{N^{2}\times C}$ along the channel dimension via an average pooling layer to obtain the one-dimensional vector $\mathbf{z}\in\mathbb{R}^{N^{2}}$.
Subsequently, a fully connected layer is further employed to transform $\mathbf{z}$ into $\mathbf{f}\in\mathbb{R}^d$, which is the input for the subsequent contrastive mechanism.

For the compactness constraint, considering that cross-batch graphs could enrich the cross-sequence context, a memory bank $\mathcal{M}$ of aggregated representations is constructed. 
The $k$-th element $\mathbf{m}_{k}\in\mathcal{M}$ denotes the class-specific aggregation that embodies unique properties of the $k$-th action class.
In practice, $\mathcal{M}$ is randomly initialized and updated using a momentum update strategy.
Given an input batch of samples, $\mathbf{m}_{k}$ is updated by
\begin{equation}
  \mathbf{m}_{k} = \alpha \cdot \mathbf{m}_{k} + (1-\alpha) \cdot \overline{\mathbf{f}_{k}}
  \label{eq_8}
\end{equation}
where $\overline{\mathbf{f}_{k}}$ is the average of $\{\mathbf{f}_{k,j}\}_{j=1}^{n_k}$, \ie, the $n_k$ samples of class $k$ within input batch, and $\alpha$ is the momentum term.
Along with the process, $\mathbf{m}_{k}$ gradually becomes a stable estimation of the clustering center for class $k$.

Then these class-specific aggregations are further optimized.
Given the feature embedding $\mathbf{f}_{k,j}$ of the $k$-th class ($j$ is the batch index), the class-specific contrastive loss $\mathcal{L}_{CSC}$ could be formulated as
\begin{equation}
  \mathcal{L}_{CSC} = -\log \frac{\exp(\mathbf{f}_{k,j} \cdot \mathbf{m}_{k} / \tau)}{\exp(\mathbf{f}_{k,j} \cdot \mathbf{m}_{k} / \tau) + \sum_{l\neq k}  \exp(\mathbf{f}_{k,j} \cdot \mathbf{m}_{l} / \tau)} 
  \label{eq_9}
\end{equation}
where $\mathbf{m}_{k}$, $\mathbf{m}_{l}$ denotes the positive and negative class aggregation \wrt $\mathbf{f}_{k,j}$, and $\tau$ is the temperature parameter.

Finally, the overall training objective function of ProtoGCN could be written as
\begin{equation}
  \mathcal{L} = \mathcal{L}_{CE} + \lambda \mathcal{L}_{CSC}
  \label{eq_10}
\end{equation}
where $\mathcal{L}_{CE}$ is the cross-entropy loss used to supervise the predicted action class, and $\lambda$ is the balance parameter.

\begin{table*}[t]
  \caption{ 
  Performance comparisons against state-of-the-art methods on the NTU RGB+D, NTU RGB+D 120, and  Kinetics-Skeleton datasets in terms of classification accuracy (\%).
  }
  \label{tab:main}
  \centering
  \setlength\tabcolsep{10pt}
  \begin{tabular}{l|c|c c|c c|c c}
    \toprule
    \multirow{2}{*}{Methods} & \multirow{2}{*}{Publication} 
    & \multicolumn{2}{c|}{NTU RGB+D} 
    & \multicolumn{2}{c|}{NTU RGB+D 120}
    & \multicolumn{2}{c}{Kinetics-Skeleton}
    \cr
    & & X-Sub & X-View & \, X-Sub & X-Set & \, Top-1 & Top-5 \cr
    \midrule
    ST-GCN \cite{yan2018spatial} & AAAI 2018 & 81.5 &  88.3 & 70.7 & 73.2 & 30.7 & 52.8 \\
    AS-GCN \cite{li2019actional} & CVPR 2019 & 86.8 &  94.2 & 78.3 & 79.8 & 34.8 & 56.5 \\
	  2s-AGCN \cite{shi2019two} & CVPR 2019 & 88.5 & 95.1 & 82.5 & 84.2 & 36.1 & 58.7 \\
    MS-G3D \cite{liu2020disentangling} & CVPR 2020 & 91.5 & 96.2 & 86.9 & 88.4 & 38.0 & 60.9 \\
	  DC-GCN+ADG \cite{cheng2020decoupling} & ECCV 2020 & 90.8 & 96.6 & 86.5 & 88.1 & - & - \\
    MST-GCN \cite{chen2021multi} & AAAI 2021 & 91.5 & 96.6 & 87.5 & 88.8 & 38.1 & 60.8 \\
    CTR-GCN \cite{chen2021channel} & ICCV 2021 & 92.4 & 96.8 & 88.9 & 90.6 & - & - \\
    STF \cite{ke2022towards} & AAAI 2022 & 92.5 & 96.9 & 88.9 & 89.9 & 39.9 & - \\
    InfoGCN \cite{chi2022infogcn} & CVPR 2022 & 93.0 & 97.1 & 89.8 & 91.2 & - & - \\
    PYSKL \cite{duan2022pyskl} & ACM MM 2022 & 92.6 & 97.4 & 88.6 & 90.8 & 49.1 & - \\
    SkeletonGCL \cite{huang2023graph} & ICLR 2023 & 92.8 & 97.1 & 89.8 & 91.2 & - & - \\
    FR-Head \cite{zhou2023learning} & CVPR 2023 & 92.8 & 96.8 & 89.5 & 90.9 & - & - \\
    GAP \cite{xiang2023generative} & ICCV 2023 & 92.9 & 97.0 & 89.9 & 91.1 & - & - \\
	  HD-GCN \cite{lee2023hierarchically} & ICCV 2023 & 93.4 & 97.2 & 90.1 & 91.6 & 40.9 & 63.5 \\
    JT-GraphFormer \cite{zheng2024spatio} & AAAI 2024 & 93.4 & 97.5 & 89.9 & 91.7 & - & - \\
    DS-GCN \cite{xie2024dynamic} & AAAI 2024 & 93.1 & 97.5 & 89.2 & 91.1 & 50.6 & - \\
	  BlockGCN \cite{zhou2024blockgcn} & CVPR 2024 & 93.1 & 97.0 & 90.3 & 91.5 & - & - \\
    \midrule
    ProtoGCN (2-ensemble) &  & 93.0 & 97.2 & 89.7 & 91.2 & 49.9 & 74.0 \\
    ProtoGCN (4-ensemble) &  & 93.5 & 97.5 & 90.4 & 91.9 & 51.3 & 75.1 \\
    \textbf{ProtoGCN (6-ensemble)} &  & \cellcolor{cyan!20}\textbf{93.8} & \cellcolor{cyan!20}\textbf{97.8} & \cellcolor{cyan!20}\textbf{90.9} & \cellcolor{cyan!20}\textbf{92.2} & \cellcolor{cyan!20}\textbf{51.9} & \cellcolor{cyan!20}\textbf{75.6} \\
    \bottomrule
  \end{tabular}
\end{table*}

\subsection{Discussion}
\label{sec:3_5}

The learning mechanism of our method is to make the network adaptively discover and assemble the learnable prototypes to generate more discriminative representations.
Accordingly, enhancing $\mathbf{X}$ in \cref{sec:3_3} and optimizing $\mathbf{Z}$ in \cref{sec:3_4} could further facilitate this process.
ProtoGCN especially improves classification for similar classes because it focuses more on specific characteristics to reveal the subtle differences of key body parts, which are essential for distinguishing these classes.

Notably, the expression employed in \cref{eq_4} constrains the model to craft representations solely using learned prototypes stored in memory module.
Driven by the contrastive learning objective, these representations are required to be discriminative, thereby amplifying the inter-class distinction in the feature space. 
Consequently, the building blocks, or prototypes, are expected to capture the characteristics of representative action units. 
After obtaining such principal components, the prototype reconstruction process naturally filters out motion patterns that are less discriminative, thereby suppressing noise and irrelevant information.
Instead, the key action details are highlighted and revealed.

\section{Experiments}

\subsection{Datasets}

\textbf{NTU RGB+D (NTU-60)}~\cite{shahroudy2016ntu} is a large-scale human action recognition dataset containing 56,880 skeleton action sequences, which are performed by 40 distinct subjects and classified into 60 classes.
Each sequence is annotated with 25 body joints and guaranteed to have at most 2 subjects.
This dataset recommends two benchmarks: 
(1) cross-subject (X-Sub): training data comes from 20 subjects, and testing data comes from other 20 subjects. 
(2) cross-view (X-View): training data comes from camera views 2 and 3, and testing data comes from camera view 1.

\noindent
\textbf{NTU RGB+D 120 (NTU-120)}~\cite{liu2019ntu} is an extended version of NTU RGB+D, and contains 114,480 videos of 120 action classes.
All skeleton sequences are performed by 106 subjects and captured from 32 different camera setups.
Similarly, the two recommended settings are suggested: 
(1) cross-subject (X-Sub): training data comes from 53 subjects, and testing data comes from other 53 subjects. 
(2) cross-setup (X-Set): training data comes from 16 even setup IDs,  and testing data comes from 16 odd setup IDs.

\noindent
\textbf{Kinetics-Skeleton (Kinetics)}~\cite{kay2017kinetics} is derived from Kinetics 400 video dataset using the pose estimation toolbox. 
It contains 240,436 training and 19,796 evaluation skeleton clips across 400 classes.
At each time step, two people are selected for multi-person clips based on the average joint confidence.
Regarding the experiments, we adopt the available skeletons provided by PYSKL~\cite{duan2022pyskl}.
Top-1 and Top-5 accuracies are used in the evaluation protocol.

\noindent
\textbf{FineGYM}~\cite{shao2020finegym} is a recent large-scale fine-grained action recognition dataset with 29,000 videos of 99 gymnastic action classes, which requires action recognition methods to distinguish different sub-actions within the same video.
We use the skeleton data provided by PYSKL~\cite{duan2022pyskl}.
The mean class Top-1 accuracy is reported in the evaluation protocol.

\subsection{Implementation Details}

In our experiments, ProtoGCN is trained for 150 epochs with the batch size set to 64. 
We employ PYSKL~\cite{duan2022pyskl} as the baseline. 
The initial learning rate is set to 0.1, and we decay it using a cosine learning rate scheduler. 
The SGD optimizer is employed with a Nesterov momentum of 0.9 and a weight decay of $5\times10^{-4}$. 
Following generic settings, momentum $\alpha$ is set to 0.9, and temperature $\tau$ is set to 0.125.
$K$ is set to 8, and the dimension $d$ for contrastive learning is set to 256.
The learnable matrices are randomly initialized for topology modeling.
We follow the data pre-processing procedures outlined in PYSKL~\cite{duan2022pyskl}. 
All experiments are conducted on a single RTX 3090 GPU with PyTorch.

\begin{table}[t]
  \caption{ 
  Performance comparisons against state-of-the-art methods on FineGYM in terms of classification accuracy (\%).
  }
  \label{tab:FineGYM}
  \centering
  \setlength\tabcolsep{9pt}
  \resizebox{\linewidth}{!}{
  \begin{tabular}{l c c}
  \toprule
  Methods & Modality & FineGYM \cr
  \midrule
  I3D \cite{carreira2017quo} & RGB & 64.4 \\
  TSN \cite{wang2016temporal} & RGB+Flow & 79.8 \\
  TSM \cite{lin2019tsm} & RGB+Flow & 81.2 \\
  LT-S3D \cite{xie2018rethinking} & RGB & 88.9 \\
  SkeletonMAE \cite{yan2023skeletonmae} & Skeleton & 91.8 \\
  PYSKL \cite{duan2022pyskl} & Skeleton & 94.1 \\
  PoseConv3D \cite{duan2022revisiting} & Skeleton+Limb & 94.3 \\
  \midrule
  \textbf{ProtoGCN (Ours)} & Skeleton & \cellcolor{cyan!20}\textbf{95.9} \\
  \bottomrule
  \end{tabular}
  }
\end{table}

\begin{figure}[t]
  \centering
  \begin{minipage}{.465\linewidth}
    \centering
    \includegraphics[width=\linewidth]{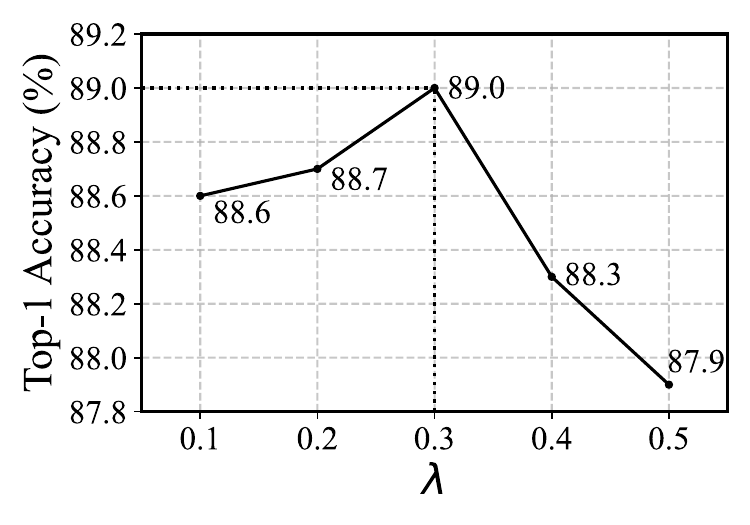}
  \end{minipage}
  \hfill
  \begin{minipage}{.515\linewidth}
    \centering
    \includegraphics[width=\linewidth]{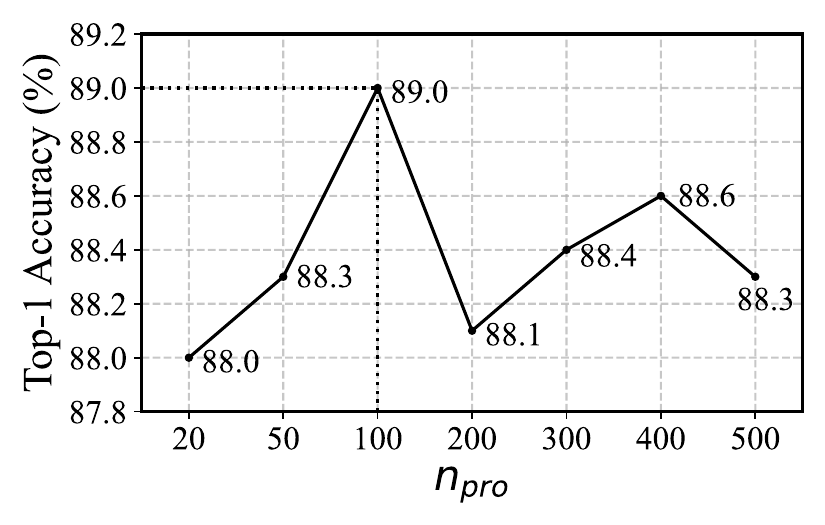}
  \end{minipage}
  \vspace{-6pt}
  \caption{
  Ablation study on the influences of weight $\lambda$ and the memory capacity $n_{pro}$ under the NTU-120 X-Sub setting.
  }
  \label{fig:figure3}
  \vspace{-2mm}
\end{figure}

\subsection{Comparison with State-of-the-Art Methods}

To improve recognition accuracy, recent state-of-the-art methods~\cite{shi2019two,chen2021channel,chi2022infogcn,lee2023hierarchically} utilize the multi-stream fusion framework. 
For a fair comparison, we adopt the widely adopted six-stream ensemble strategy proposed in InfoGCN~\cite{chi2022infogcn}.

We compare ProtoGCN with extensive benchmark methods on four datasets, including NTU-60, NTU-120, Kinetics, and FineGYM.
According to the results reported in~\cref{tab:main} and~\cref{tab:FineGYM}, ProtoGCN consistently achieves the best accuracy and outperforms other approaches by a significant margin in all scenarios. 
On the NTU-60 dataset, the proposed method surpasses the performance of BlockGCN~\cite{zhou2024blockgcn} by 0.7\% on X-Sub and 0.8\% on X-View.
Remarkably, ProtoGCN excels in performance on the challenging NTU-120 dataset, achieving accuracies of 90.9\% on X-Sub and 92.2\% on X-Set.
As for Kinetics, ProtoGCN outperforms state-of-the-art method DS-GCN~\cite{xie2024dynamic} by 1.3\%. 
According to \cref{tab:FineGYM}, ProtoGCN has achieved a significant improvement in fine-grained action recognition, further confirming the efficacy of our approach.

\begin{table}[t]
  \caption{
  Ablation study on the contribution of each component in ProtoGCN under the NTU-120 X-Sub setting.
  }
  \label{tab:each}
  \centering
  \setlength\tabcolsep{9pt}
  \begin{tabular}{c c c c c}
  \toprule
  Baseline & MTE & PRN & CSCL & Acc (\%) \cr
  \midrule
  \checkmark & -- & -- & -- & 87.8 \\
  \checkmark & \checkmark & -- & -- & 88.1 ($\uparrow$ 0.3) \\
  \checkmark & \checkmark & -- & \checkmark & 88.3 ($\uparrow$ 0.5) \\
  \checkmark & \checkmark & \checkmark & -- & 88.5 ($\uparrow$ 0.7) \\
  \checkmark & \checkmark & \checkmark & \checkmark & \cellcolor{cyan!20}\textbf{89.0 ($\uparrow$ 1.2)} \\
  \bottomrule
  \end{tabular}
\end{table}

\begin{table}[t]
  \caption{
  Comparison of classification accuracies (\%) based on (\textit{left}) existing graph contrastive methods and (\textit{right}) different enhanced topologies for the network.
  }
  \label{tab:graph}
  \centering
  \begin{minipage}{.51\linewidth}
    \centering
    \setlength\tabcolsep{0pt}
    \begin{tabular}{l c}
      \toprule
      Methods & Acc (\%) \cr
      \midrule
      Baseline & 87.8 \\
      w/ SkeletonGCL \cite{huang2023graph} & 88.0 \\
      w/ FR-Head \cite{zhou2023learning} & 88.2 \\
      Ours & \cellcolor{cyan!20}\textbf{89.0} \\
      \bottomrule
    \end{tabular}
  \end{minipage}
  \hfill
  \begin{minipage}{.47\linewidth}
    \centering
    \setlength\tabcolsep{0pt}
    \begin{tabular}{l c}
      \toprule
      Methods & Acc (\%) \cr
      \midrule
      ProtoGCN & \cellcolor{cyan!20}\textbf{89.0} \\
      w/o $\mathbf{A_{intra}}$ & 88.8 \\
      w/o $\mathbf{A_{inter}}$ & 88.5 \\
      w/o $\mathbf{A_{intra}}, \mathbf{A_{inter}}$ & 88.4 \\
      \bottomrule
    \end{tabular}
  \end{minipage}
\end{table}

\begin{table}[t]
  \caption{ 
  Average performance (\%) on different difficulty level classes sorted by accuracy under the NTU-120 X-Sub setting.
  }
  \label{tab:average}
  \centering
  \setlength\tabcolsep{9.8pt}
  \begin{tabular}{l c c c}
  \toprule
  Difficulty Level  & Baseline & ProtoGCN & $\Delta$ \\
  \midrule
  class 1-10 & 64.1 & 67.1 & \cellcolor{cyan!20}\textbf{+3.0} \\
  class 11-30 & 78.6 & 81.1 & \cellcolor{cyan!20}\textbf{+2.5} \\
  class 31-60 & 88.6 & 89.7 & \cellcolor{cyan!20}\textbf{+1.1} \\
  class 61-120 & 96.0 & 96.4 & \cellcolor{cyan!20}\textbf{+0.4} \\
  \bottomrule
  \end{tabular}
  \vspace{-2mm}
\end{table}

\subsection{Ablation Study}

In this section, we conduct an experimental evaluation to assess the effectiveness of each component in the proposed method. 
The experiments are conducted under the X-Sub setting of NTU-120 with the bone modality.

\begin{figure*}[t]
\centering
\includegraphics[width=\linewidth]{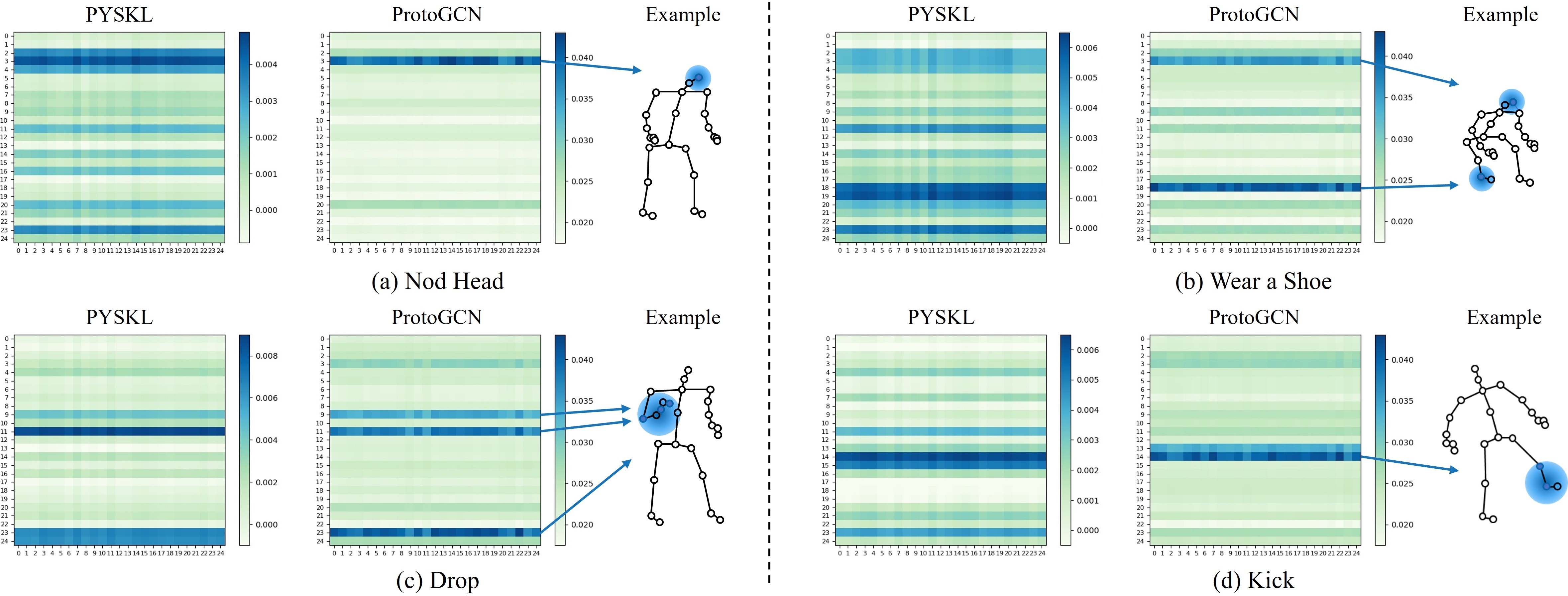}
\setlength{\abovecaptionskip}{0pt}
\setlength{\belowcaptionskip}{0pt}
\vspace{-0.3cm}
\caption{ 
Visualization of the topologies learned by PYSKL~\cite{duan2022pyskl} and ProtoGCN across four actions. 
Darker color indicates stronger correlation between corresponding joints.
}
\label{fig:figure4}
\vspace{-5mm}
\end{figure*}

\begin{figure}[t]
\centering
\includegraphics[width=\linewidth]{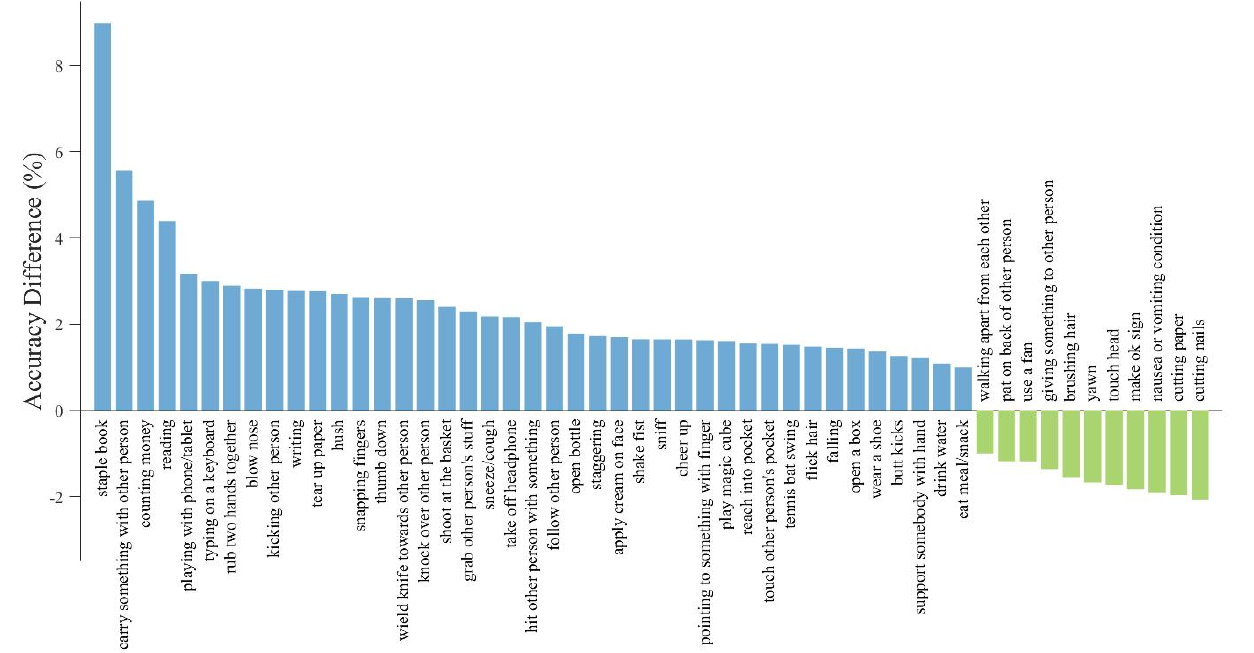}
\setlength{\abovecaptionskip}{-8pt}
\setlength{\belowcaptionskip}{0pt}
\caption{
Action classes with accuracy difference higher than 1\% between our method and PYSKL~\cite{duan2022pyskl} on the NTU-120 dataset.
}
\label{fig:figure5}
\vspace{-4mm}
\end{figure}

\noindent
\textbf{Effectiveness of Each Module in ProtoGCN.}
We examine the contribution of each component of ProtoGCN as shown in \cref{tab:each}.
The baseline model is re-trained using the latest officially released code and achieved a performance of 87.8\%.
Subsequently, we observe a performance improvement of 0.3\% with the Motion Topology Enhancement (MTE) module.
Building upon the foundation provided by MTE, $\mathcal{L}_{CSC}$ could improve performance by 0.2\%.
Interestingly, the performance significantly improves by 0.7\% (from 88.3\% to 89.0\%) with the addition of Prototype Reconstruction Network (PRN).
This indicates that PRN helps to capture the discriminative semantics through discovering diverse prototypes, thereby offering greater advantages for graph convolution learning.
By integrating all components, we surpass the baseline model by 1.2\%, demonstrating the effectiveness of the modules introduced in this work.

\noindent
\textbf{Effectiveness of Prototype Reconstruction.}
To underscore the efficacy of the reconstruction mechanism, we conduct the analysis between ProtoGCN and existing methods that employ graph contrastive objective, including SkeletonGCL~\cite{huang2023graph} and FR-Head~\cite{zhou2023learning}.
As shown in \cref{tab:graph}, while these methods facilitate the differentiation, the improvement of ProtoGCN is remarkable (1.0\% / 0.8\%).
This indicates that the mechanism of expressing features as the combinations of prototypes could further boost the accuracy.

\noindent
\textbf{Influences of Topology Decomposition.}
We assess the influences of topology decomposition in the MTE module.
We closely examine the effects of these terms and present their performance in \cref{tab:graph}.
The results demonstrate that intricate motion details are advantageous for recognition, and removing them will detrimentally affect performance.

\noindent
\textbf{Effect of Hyperparameters.}
We examine the effect of weight $\lambda$ and memory capacity $n_{pro}$ related to $\mathbf{W_{memory}}$ in \cref{fig:figure3}.
The optimal value for $\lambda$ is determined to be 0.3, corresponding to the peak accuracy.
In addition, we explore the impact of varying $n_{pro}$.
Intuitively, the memory capacity is closely related to representative patterns among all the joints.
Through empirical investigation, we determine the optimal balance and set $n_{pro}$ to 100 for the performance.

\noindent
\textbf{Performance on Similar Classes.}
We conduct the analysis of the performance in discerning actions with differing degrees of similarity. 
To achieve this, we sorted 120 classes of  NTU-120 based on accuracies, ranging from low to high according to the results obtained with the baseline. 
Subsequently, we categorized them into four difficulty levels and computed the average accuracy for each level. 
As shown in \cref{tab:average}, ProtoGCN yields a more significant accuracy gain, especially in classes with higher difficulties.
In \cref{fig:figure5}, we further present the action classes that exhibit over 1\% absolute accuracy difference with the baseline on the NTU-120 dataset.
It is evident that ProtoGCN offers more pronounced improvements in a significantly greater number of classes.

\noindent
\textbf{Visualization.}
To visually demonstrate the effectiveness of ProtoGCN, in \cref{fig:figure4}, we present the topologies for four different action classes.
Each learned prototype is a pattern vector thus it would be difficult to understand its effect through the direct individual visualization. 
Instead, we visualize the combined prototypes, \ie, the reconstructed topologies, which are shown by averaging $\mathbb{R}^{N\times N\times C}$ along the $C$ dimension.
By comparing the results of ProtoGCN with the baseline PYSKL, our method showcases an ability to effectively focus on the joints most relevant to the target motion. 
Furthermore, a closer inspection reveals that the proposed method can further accentuate the key motion details, such as color patterns or intensity fluctuations within the darkened rows.
These fine-grained joint-level relationships are crucial for distinguishing similar actions.

\section{Conclusion}

In this paper, we propose ProtoGCN, a novel GCN-based method for skeleton-based action recognition.
ProtoGCN highlights its prototype reconstruction mechanism, aimed at revealing the fine-grained motion patterns essential for distinguishing similar actions. 
Through the addition of enhanced topology representations, ProtoGCN is empowered by the contrastive objective to learn and identify distinctive action units.  
Consequently, the reconstructed motion feature becomes sufficiently discernible, enabling the discrimination of similar actions.
The effectiveness of ProtoGCN is substantiated by achieving state-of-the-art performance across four widely utilized benchmark datasets.

\section*{Acknowledgments}

This work is supported by the National Natural Science Foundation of China (Grant No.U23B2054, 62276263, 62406304, and 62406028), and Tianjin Key Research and Development Program CAS-Cooperation Project (Grant No.24YFYSHZ00290).

{
    \small
    \bibliographystyle{ieeenat_fullname}
    \bibliography{main}
}

\end{document}